\documentclass{article} 

\usepackage[]{log_2025}
\usepackage{booktabs}						
\usepackage{multirow}						
\usepackage{amsfonts}						
\usepackage{graphicx}						
\usepackage{duckuments}	

\usepackage{url}
\usepackage{amsmath}
\usepackage{subcaption}
\usepackage{xcolor}   
\usepackage{colortbl} 
\usepackage{bm}
\usepackage{amsfonts}
\usepackage{multirow}
\usepackage{multicol}
\usepackage[numbers,compress,sort]{natbib}
\usepackage{mathtools} 
\usepackage{booktabs} 
\usepackage{tikz} 
\usepackage{paralist} 
\usepackage{amsthm}
\usepackage{wrapfig}

\newtheorem{theorem}{Theorem}[section] 
\newtheorem{lemma}[theorem]{Lemma}
\newtheorem{definition}{Definition}

\definecolor{mydarkblue}{rgb}{0,0.08,0.45}
\definecolor{myblue}{HTML}{3b75c3}
\definecolor{myred}{HTML}{E33222}
\definecolor{mygreen}{HTML}{438773}
\definecolor{mymaroon}{RGB}{142,27,19}
\definecolor{maroon}{HTML}{800000}
\definecolor{mycite}{cmyk}{0.55,1,0,0.15}
\definecolor{codeblue}{rgb}{0.25,0.5,0.5}
\definecolor{codekw}{rgb}{0.85, 0.18, 0.50}
\definecolor{codegreen}{rgb}{0,0.6,0}
\definecolor{codegray}{rgb}{0.5,0.5,0.5}
\definecolor{codepurple}{rgb}{0.58,0,0.82}
\definecolor{backcolour}{rgb}{0.95,0.95,0.92}

\newcommand{\showcomment}{} 
\newcommand{\addcomment}[2]{\ifdefmacro{\showcomment}{{\textcolor{#1}{#2}}}{}}

\newcommand{\ns}[1]{{\addcomment{purple}{[Neil: #1]}}}

\newcommand{\sz}[1]{{\addcomment{orange}{[SZ: #1]}}}

\newcommand{\nop}[1]{}
\newcommand{\method}[0]{EHDM}
\title{
Weak Models Can be Good Teachers: A Case Study on Link Prediction with MLPs
}

\author[Qin et al.]{%
  Zongyue Qin\textsuperscript{1}
  \And Shichang Zhang\textsuperscript{2}
  \And Mingxuan Ju\textsuperscript{3}
  \AND 
  Tong Zhao\textsuperscript{3}
  \And Neil Shah\textsuperscript{3}
  \And Yizhou Sun\textsuperscript{1}\\\AND
  \ \vspace{-6mm}\\
  \textsuperscript{1}University of California Los Angeles, Los Angeles, California, USA\\
  \textsuperscript{2}Harvard University, Boston, Massachusetts, USA\\
  \textsuperscript{3}\thanks{Authors affliated with Snap Inc. served in advisory roles only for this work.}\ \ Snap Inc., Bellevue, Washington, USA
}

\begin{document}
\maketitle

\begin{abstract}

Link prediction is a crucial graph-learning task. Distilling Graph Neural Network (GNN) teachers into Multi-Layer Perceptron (MLP) students has emerged as an effective approach to achieve strong performance and reducing computational cost by removing graph dependency in the inference stage, especially in applications such as citation prediction and product recommendation where node features are abundant. However, existing distillation methods only use standard GNNs.  Do stronger models such as those specially designed for link prediction (e.g., GNN4LP) lead to better students? Are heuristic-based methods (e.g., common neighbors) bad teachers as they are weak models? \nop{and overlook alternative teachers such as specialized model for link prediction (GNN4LP) and heuristic methods (e.g., common neighbors). }
This paper first explores the impact of different teachers in MLP distillation. Surprisingly, we find that stronger models do not always produce stronger students: MLPs distilled from GNN4LP can underperform those distilled from simpler GNNs, while weaker heuristic methods can teach MLPs to near-GNN performance with drastically reduced training costs. We provide both theoretical and empirical analysis to explain this phenomenon, revealing that a teacher is only as good as its \emph{teachable knowledge}, the portion of its knowledge that can be transferred through the features accessible to the student. 
Building on these insights, we propose Ensemble Heuristic-Distilled MLPs (\method{}), which eliminates costly GNN training while effectively training complementary MLP predictors via different heuristic teachers. Our extensive experiments show \method{} reduces the total training time by \textbf{1.95-3.32$\times$} while achieve an average \textbf{7.93\%} improvement over previous GNN-to-MLP approaches, indicating that it is an efficient and effective link prediction method.

\end{abstract}

\section{Introduction}\label{sec:intro}


Link prediction is a pivotal task in graph learning, with widespread usage in web applications. It involves estimating the likelihood of a link between two nodes, enabling applications including citation prediction, friend recommendation, knowledge graph completion~\citep{schlichtkrull2018modeling,nathani2019learning,vashishth2019composition}, and item recommendation~\citep{koren2009matrix,ying2018graph,he2020lightgcn}. 
Graph neural networks (GNNs)~\citep{kipf2016semi,hamilton2017inductive,velivckovic2017graph} have been widely used for link prediction. This includes standard generic GNNs~\citep{kipf2016variational,davidson2018hyperspherical}, as well as specifically designed GNNs for link prediction (GNN4LP) which are augmented with additional structural features~\citep{zhang2018link,zhang2021graph,zhu2021neural,yun2021neo,ncnc,linkmoe} to capture extra link-specific information~\citep{zhang2021labeling}. However, all of these GNN models suffer from heavy computations in inference time due to their dependency on graphs, which require repeatedly fetching and aggregating neighboring nodes for GNN message passing~\citep{zhang2021graph}.

Researchers have proposed various methods to accelerate GNN inference. There are pruning~\citep{zhou2021accelerating,chen2021unified} and quantization~\citep{zhao2020learned,tailor2020degree} methods that accelerate GNN inference to some extent. But the speed-up is limited because they do not resolve the graph dependency issue caused by the message passing. An alternative line of work focuses on distilling knowledge of GNN teachers into Multi-Layer Perceptron (MLP) students that only take node features as inputs~\citep{zhang2021graph,tian2022learning,wu2023extracting,wu2023quantifying,wu2024teach,lu2024adagmlp,wang2025training}. By eliminating graph dependency and expensive neighborhood aggregation, MLPs offer up to 70-273$\times$ faster inference~\citep{guo2023linkless,zhang2021graph}. Meanwhile, GNN-distilled MLPs can match or even surpass the performance of GNN teachers in many cases where rich node features are available (e.g., recommendation systems).

Advances in MLP distillation on graphs focus primarily on node classification with only a few works extending the idea to link prediction. For example, LLP~\citep{guo2023linkless} distills relational knowledge from standard GNNs to MLPs to handle link prediction. Yet, the link-level GNN-to-MLP distillation remains under-explored. Besides, existing MLP distillation works rely majorly on standard GNNs as teachers. In reality, state-of-the-art link prediction models often go beyond standard GNNs. GNN4LP methods, or even classical heuristic methods (e.g., common neighbors)~\citep{kumar2020link,newman2001clustering,murase2019structural,zhou2009predicting}, were shown to outperform standard GNNs in many scenarios~\citep{heart}. This creates an intriguing question: \emph{How does the choice of teacher model influence the MLP student’s performance?}

Moreover, although GNN-to-MLP distillation accelerates inference, it still requires training a GNN teacher, which is itself computationally expensive in large-scale applications. Fine-tuning hyper-parameters for GNNs further inflates this cost. Hence, there is a growing incentive to identify alternative teachers that can distill useful link-level information at lower training cost.


In this work, we first explore how different teachers, ranging from GNNs to GNN4LP methods and heuristic-based approaches, affect the performance of the distilled MLP student for link prediction. Adopting the LLP framework~\citep{guo2023linkless} as the distillation pipeline, we have two surprising observations:
\begin{compactenum}[\textbullet]
    \item \emph{Good models $\neq$ good teachers.} Although GNN4LP models often outperform standard GNNs, the MLPs distilled from GNN4LP can actually underperform those distilled from simpler GNNs. 
    \item \emph{Heuristic methods can be good teachers, even when they underperform.} Classical heuristics can be effective teachers for MLPs to guide them achieve comparable performance to standard GNNs, even if the heuristic’s performance is worse than the MLP. 
\end{compactenum}

To explain these findings, we provide both empirical and theoretical analyses. The first observation arises because MLPs, with their limited capacity and lack of message passing, struggle to absorb the intricate structural patterns that GNN4LP teachers rely on. In contrast, the second observation highlights that heuristic methods, despite their simplicity and weaker standalone performance, offer complementary signals that are easier for MLPs to learn and align with. 

Based on these insights, we propose to distill MLPs from heuristic teachers, which will substantially shorten overall training time.
To further improve prediction accuracy, we design a simple yet effective Ensemble Heuristic-Distilled MLPs (\method{}) method that trains one MLP per heuristic, then learns a gater MLP to fuse their predictions at inference time. Our gating mechanism uses \emph{only} node features, thereby preserving the MLP’s fast, neighbor-free inference. This allows us to harness multiple heuristics’ complementary signals and achieve higher accuracy without reintroducing graph dependencies. 
To summarize, this work makes the following contributions:
\begin{compactitem}[\textbullet]
    \item We show that even underperforming heuristic methods can be surprisingly good teachers to MLPs, substantially reducing the training cost while enabling near-GNN predictive performance.
    \item We provide empirical and theoretical analysis to this phenomenon, revealing that a teacher is only as good as its \emph{teachable knowledge}, the portion of its knowledge that can be transferred through the features accessible to the student.
    \item Our work is the first to formally prove that a more expressive model class (GNN4LP) does not inherently offer better teachable knowledge than a simpler model class (GNN). This result highlights an important lesson for future GNN2MLP research: rather than focusing on teacher strength, we should aim to design models that are more teachable.
    \item We propose an ensemble method \method{} to efficiently and effectively train multiple MLPs that capture different positive links under different heuristic guidance. Experiments on ten datasets show a reduction of training time by \textbf{1.95-3.32$\times$}, with an average of \textbf{7.93\%} performance gain and comparable inference speed over previous GNN-to-MLP methods~\citep{guo2023linkless}. 
\end{compactitem}

\section{Preliminaries and Related Work}\label{sec:prelim}


\subsection{Link Prediction on Graphs}

Let $G=(\mathcal{V}, \mathcal{E})$ be an undirected graph with $N$ nodes and observed edges $\mathcal{E}\subset \mathcal{V}\times\mathcal{V}$. Node features are represented as $\mathbf{X}\in\mathbb{R}^{N\times F}$ where $\bm{x}_i$ is the feature vector for node $i$. Given a pair of nodes $(i,j)$, the goal of link prediction is to estimate the probability of connection, denoted as $P(Y_{ij}=1)$. 

\paragraph{Graph Neural Network (GNN) Solutions} The state-of-the-art solutions are based on GNNs. Early methods directly utilize the embeddings of two target nodes for prediction~\citep{kipf2016variational,berg2017graph,schlichtkrull2018modeling,ying2018graph,davidson2018hyperspherical}. Subsequent research revealed that standard GNN methods struggle to capture link-specific information~\citep{zhang2021labeling}. To overcome this limitation, \textbf{GNN4LP models} extend GNNs by explicitly incorporating structural information that is not inherently encoded by GNN architectures~\citep{heart}. A common example is the inclusion of global node IDs, since vanilla GNN inputs typically contain only node features. These IDs can then be exploited to compute structural signals such as the number of common neighbors~\citep{yun2021neo,ncnc} and subgraph features~\citep{zhang2021labeling,buddy,nbfnet,mplp}.
Although these methods improve prediction accuracy, they increase the computational cost during training and inference, rendering them less suitable for large-scale web applications.

\paragraph{Heuristic Solutions} Heuristic methods estimate node proximity based on structural or feature similarities~\citep{mao2023revisiting}, categorized as follows:

\begin{compactitem}[\textbullet]
    \item \textbf{Local Structural Proximity} considers 1-hop neighborhood similarity. We consider three heuristics: \text{Common Neighbors (CN)}~\citep{newman2001clustering}:  
        $
        CN(i,j) = |N(i) \cap N(j)|
        $, 
       \text{Adamic-Adar (AA)}~\citep{adamic2003friends}:  
        $
        AA(i,j) = \sum_{k \in N(i) \cap N(j)} \frac{1}{\log(\text{Deg}(k))}
        $,
        \text{Resource Allocation (RA)}~\citep{zhou2009predicting}:  
        $
        RA(i,j) = \sum_{k \in N(i) \cap N(j)} \frac{1}{\text{Deg}(k)}
        $, where $N(i)$ is the neighbors of node $i$ and $\text{Deg}(\cdot)$ is the node degree.

    \item \textbf{Global Structural Proximity} captures higher-order connectivity, e.g., SimRank~\citep{jeh2002simrank}, Katz~\citep{katz1953new}, and Personalized PageRank~\citep{brin1998anatomy}. Due to their high computational cost, we adopt \text{Capped Shortest Path (CSP)}, a simplified shortest-path heuristic~\citep{liben2003link}:  
    $
    CSP(i,j) = \frac{1}{\min(\tau, SP(i,j))}
    $,
    where \( SP(i,j) \) is the shortest path length between \( i \) and \( j \), and \( \tau \) limits the maximum distance considered\footnote{With bi-directional breadth first search, we need to retrieve $\tau/2$-hops neighborhood to compute CSP.}.

    \item \textbf{Feature Proximity} measures similarity in node features, assuming nodes with similar characteristics are more likely to connect~\citep{khanam2023homophily}. Ma et al.~\citep{linkmoe} shows MLP is good at capturing feature proximity.
\end{compactitem}

\subsection{GNN Acceleration}

Inference acceleration has been explored through both hardware and algorithmic optimizations. On the algorithmic side, techniques such as pruning~\citep{han2015learning,zhou2021accelerating,chen2021unified} and quantization~\citep{gupta2015deep,zhao2020learned, tailor2020degree} have been widely studied to reduce model complexity and computational cost. 
While these techniques improve inference efficiency, they do not eliminate neighbor-fetching operation, which remains a fundamental bottleneck in GNN inference. Because GNNs rely on message passing across graph structures, even optimized models still suffer from high data dependency and irregular memory access patterns. 

\subsection{GNN-to-MLP Distillation} 

To further improve inference efficiency in graph applications, GLNN~\citep{zhang2021graph} pioneered GNN-to-MLP distillation, eliminating the need for neighbor fetching and significantly accelerating inference. Notably, \textbf{the MLPs only take node features as inputs}, but they can learn to infer the structural information from the node features during distillation. Subsequent works have refined distillation techniques~\citep{wu2023extracting,wu2023quantifying,wu2024teach,lu2024adagmlp,tian2022learning}, but mainly for node classification. Wang et al.~\citep{wang2025training} introduced a self-supervised approach applicable to node classification, link prediction, and graph classification. However, their method is restricted to GNN teachers and does not generalize to GNN4LP models or heuristic methods. The first extension of GNN-to-MLP distillation for link prediction was LLP~\citep{guo2023linkless}, but like prior methods, it only explores GNN teachers.
Notably, LLP’s framework is applicable beyond standard GNNs, as it only requires prediction from the teachers. So we adopt LLP’s framework while extending it to other teacher options. The details of LLP are introduced in Appendix \ref{app:llp}.

\section{Key Observations and Theoretical Analysis}
In this section, we present key observations that motivate the use of heuristic methods to distill MLP. First, we compare GNN and GNN4LP models as teachers and find that GNN4LP, despite achieving better accuracy, does not always serve as a superior teacher. We then highlight the advantages of using heuristics as teachers: (1) they significantly enhance MLP performance and (2) they offer better training efficiency.

\subsection{Better Models $\ne$ Better Teachers}\label{sec:better_model}

\begin{wrapfigure}{r}{0.5\textwidth}
    \centering
    \begin{subfigure}{0.23\textwidth}
        \centering
        \includegraphics[width=\textwidth]{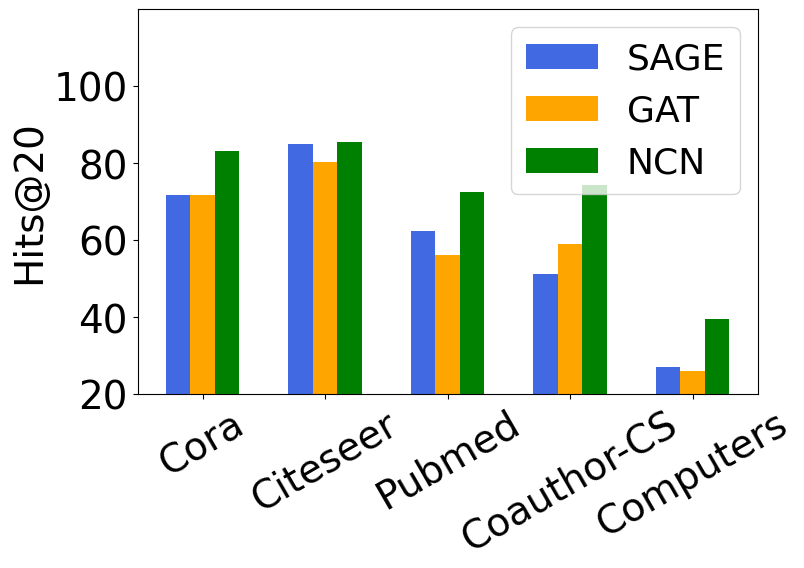}
        \caption{Hits@20 of teachers}
        \label{fig:sage_gat_ncn_teacher}
    \end{subfigure}
    \hfill
    \begin{subfigure}{0.23\textwidth}
        \centering
        \includegraphics[width=\textwidth]{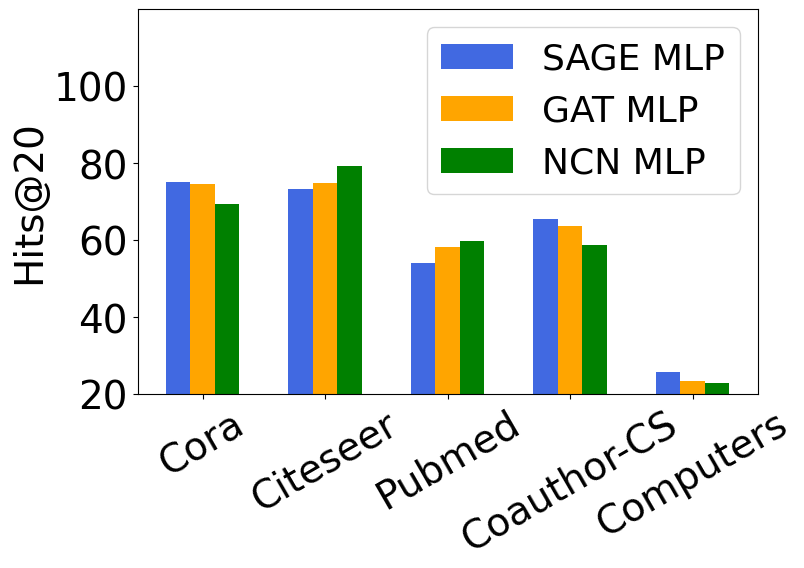}
        \caption{Hits@20 of students}
        \label{fig:sage_gat_ncn_student}
    \end{subfigure}
    \caption{Hits@20 of standard GNNs (SAGE, GAT), a GNN4LP model (NCN), and their student MLPs across five datasets.}
    \label{fig:sage_gat_ncn}
    \vspace{-3mm}
\end{wrapfigure}

Unlike node and graph classification tasks, state-of-the-art GNN4LP models for link prediction fall outside the scope of standard message-passing neural networks (MPNNs). 
While they generally outperform standard GNNs, our results surprisingly show that this advantage does not translate into better MLP students.
As shown in Figure \ref{fig:sage_gat_ncn}, while NCN~\citep{ncnc} substantially outperforms SAGE~\citep{hamilton2017inductive} and GAT~\citep{velivckovic2017graph}, NCN-distilled MLPs exhibit suboptimal performance in three out of five datasets. 

This aligns with prior observations that stronger teachers do not always produce better students~\citep{wu2024teach,jafari2021annealing,zhu2021student,qiu2022better}, often due to a capacity mismatch. GNN4LP models exploit structural information that standard GNNs cannot capture. Since MLPs rely solely on node features, they cannot inherit the performance gains of GNN4LPs when those gains come from feature-independent structural patterns.

To formalize this intuition, we introduce a concept called ``teachable knowledge'',  which captures the part of a teacher's knowledge that can be learned by a MLP limited to node features.
\begin{definition}[Teachable Knowledge]
    Let $\bm{x}_i,\bm{x}_j$ be the features of node $i,j$. Given a teacher model $F$, its teachable knowledge to a MLP student is the conditional expectation 
    $\mathbb{E}(F|\bm{x}_i,\bm{x}_j)$, which represents the component of the teacher's predictions that is explainable by node features.
\end{definition}


The teachable knowledge represents the best possible approximation of the teacher’s output by any MLP that only has access to node features under MSE or KL-divergence. Let $s_i$ be latent random variables representing the structural information of node $i$. The following Lemma formalizes the teacher can be replaced by its teachable knowledge without changing the training objective under KL-divergence. The proof is given in Appendix \ref{app:lemma}.

\begin{lemma}
\label{lemma}
Let $F(y\mid \bm{x}_i,\bm{x}_j,s_i,s_j)$ be a teacher model, and let $g(y\mid \bm{x}_i,\bm{x}_j)$ be a student model. Suppose distillation is performed using KL divergence as the loss. Then,
\begin{equation}
    \mathbb{E}_{\bm{x}_i,\bm{x}_j,s_i,s_j}KL(F,g)=\mathbb{E}_{\bm{x}_i,\bm{x}_j,s_i,s_j}KL(\mathbb{E}[F|\bm{x}_i,\bm{x}_j],g)
\end{equation}
\end{lemma}

Therefore, if two teachers have the same teachable knowledge, they yield the same distillation loss when training the student MLP, i.e., their effectiveness as teachers is theoretically equivalent in the GNN2MLP setting.

Next, we consider the teachable knowledge of GNN and GNN4LP teachers. Let us decompose the latent variable $s_i$ into two components: $s_i=(s_i^\text{GNN}, s_i^\text{extra})$, where $s_i^\text{GNN}$ contains structural information that can be captured by GNNs, which is usually $K$-hop neighborhood of node $i$\footnote{It only contains node feautres, but not node IDs}, and $s_i^\text{extra}$ is the extra structural information beyond GNNs' capacity (e.g., the unique node IDs of node $i$'s $K$-hop neighborhood\footnote{NCN uses node IDs to identify common neighbors between two target nodes}). 
GNN4LP models achieve higher accuracy than conventional GNNs precisely because they incorporate $s_i^{extra}$ into their predictions~\citep{buddy,zhang2021labeling}. However, the following theorem formalizes why $s_i^{extra}$ does not help the student MLPs.
The proof is given in Appendix \ref{app:proof}.

\begin{theorem}[GNN4LP models are not better teachers]
\label{thm:distillation-parity}
Let a standard GNN be written as $F_\text{GNN}(\bm{x}_i, \bm{x}_j, s_i^\text{GNN}, s_j^\text{GNN})$. And a GNN4LP model as $F_\text{GNN4LP}(\bm{x}_i, \bm{x}_j, s_i^\text{GNN}, s_j^\text{GNN}, s_i^\text{extra}, s_j^\text{extra})$.
For any $F_\text{GNN4LP}$, there exists a corresponding $F_\text{GNN}$ such that the teachable knowledge from either teacher is the same:
\begin{equation}
    \begin{aligned}
&\mathbb{E}_{s^\text{GNN}, s^\text{extra}} \left[ F_\text{GNN4LP} \mid \bm{x}_i, \bm{x}_j \right]
= \mathbb{E}_{s^\text{GNN}} \left[ F_\text{GNN} \mid \bm{x}_i, \bm{x}_j \right]
    \end{aligned}
\end{equation}
\end{theorem}

In other words, there always exists a GNN teacher that can impart the same information to MLP, so that its distilled MLP matches the performance of any MLP distilled from a GNN4LP teacher.



\subsection{Heuristic Methods are good teachers}\label{sec:heuristic_good}

\begin{figure*}[ht]
    \centering
    \begin{subfigure}{0.32\linewidth}
        \centering
        \includegraphics[width=\linewidth,height=0.12\textheight]{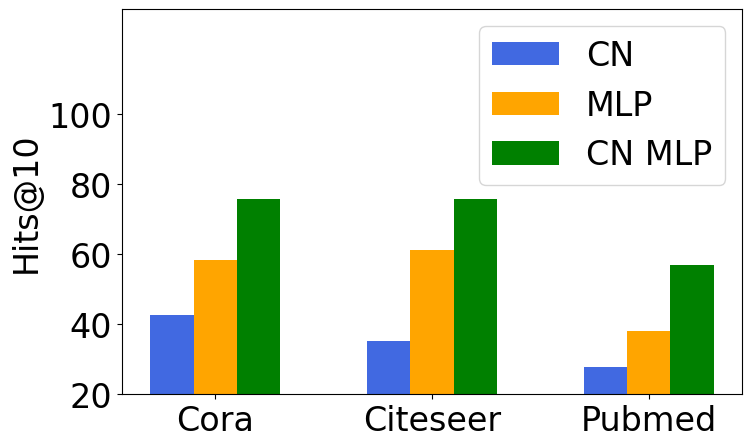}
        \caption{Hits@10 of CN, MLP and CN-distilled MLP.}
        \label{fig:CN_MLP_CNMLP}
    \end{subfigure}
    \hfill
    \begin{subfigure}{0.32\linewidth}
        \centering
        \includegraphics[width=\linewidth,height=0.1\textheight]{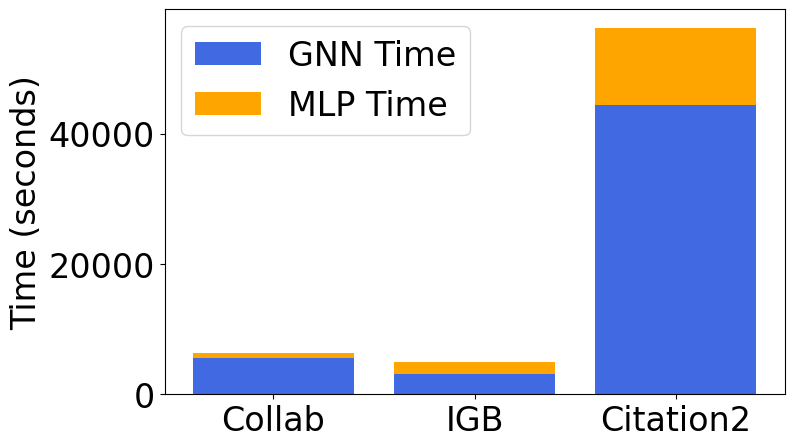}
        \caption{Time decomposition of GNN-to-MLP distillation spent on GNN and MLP.}
        \label{fig:bottleneck}
    \end{subfigure}
    \hfill
    \begin{subfigure}{0.32\linewidth}
        \centering
        \includegraphics[width=\linewidth,height=0.1\textheight]{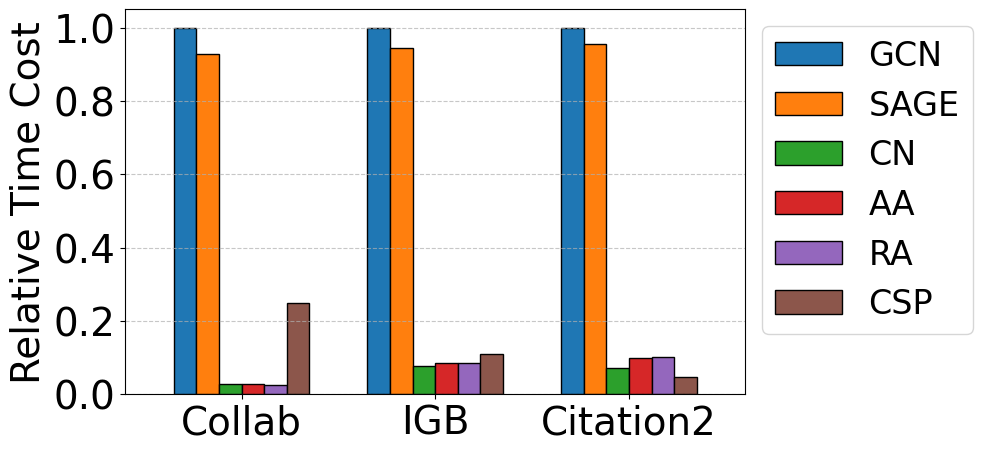}
        \caption{Relative time cost (w.r.t. GCN) to generate guidance with different teachers on large-scale datasets.}
        \label{fig:guidance_time}
    \end{subfigure}
    \caption{Exploration of heuristic methods as teachers to distill MLPs. 
    (a) Hits@10 comparison of CN, MLP, and CN-distilled MLP, demonstrating performance improvements through heuristic distillation even when the heuristic method underperforms. 
    (b) Time breakdown of GNN-to-MLP distillation, showing the computational burden of GNN training. 
    (c) Relative time cost for generating guidance with different teachers, illustrating the efficiency advantage of heuristic methods.}
    \label{fig:heuristic_advantage}
\end{figure*}

Since stronger GNN4LP models are not better teachers due to \emph{capacity mismatch}, we explore using simple heuristic methods as teachers. Unlike GNNs, heuristic methods typically capture a single type of structural information (e.g., common neighbors)~\citep{mao2023revisiting}, making them easier for distillation. 
However, their performance is highly data-dependent, and may even fall below that of MLPs. Surprisingly, we find that utilizing heuristic teachers not only offers substantial training time savings, but also produces effective students, even when heuristic methods themselves perform poorly.

\paragraph{Observation 1: Heuristic methods are much faster than GNNs.}  Despite their effectiveness, GNNs and GNN4LP models suffer from high computational costs during both training and inference. Figure \ref{fig:bottleneck} illustrates the time breakdown of GNN-to-MLP distillation on three large-scale datasets: \texttt{OGBL-Collab}, \texttt{OGBL-Citation2}~\citep{ogb}, and \texttt{IGB}~\citep{khatua2023igb}. The total GNN time includes both GNN training and guidance generation, while the MLP time refers to the MLP distillation process based on GNN guidance. We observe that 62.2\% to 87.1\% of the total time is spent on GNNs, making it the primary bottleneck in the distillation pipeline.

In contrast, heuristic methods offer a far more efficient alternative as teachers, as heuristics require no training and involve only simple calculations. Figure \ref{fig:guidance_time} compares the relative time costs of two GNN models (GCN and SAGE) against four heuristic methods. The time cost of GNN models includes both training and guidance generation, while the time cost of heuristic methods only includes guidance generation. The results indicate that using heuristic methods as teachers is an order of magnitude faster than using GNNs. 

\paragraph{Observation 2: Heuristic methods are effective teachers for MLPs, even when they underperform.} Figure~\ref{fig:CN_MLP_CNMLP} shows that CN-distilled MLPs significantly outperform non-distilled MLPs, despite CN itself having lower accuracy. This suggests that CN provides useful information that MLPs alone cannot capture. Additional results in Section \ref{sec:experiment} further confirm that heuristic-distilled MLPs achieve near-GNN performance across various datasets. These findings provide strong empirical evidence supporting the effectiveness of heuristic teachers.

This leads to a natural question: \emph{why can heuristic methods improve MLPs even when they have lower accuracy?} We argue that heuristic methods provide complementary information to MLPs. Mao et al.~\citep{mao2023revisiting} show that structural and feature proximity identify different positive node pairs, and propose the following model:
\begin{equation}
   P(Y_{ij}=1)= \left\{\begin{aligned}
     \frac{1}{1+e^{\alpha(d_{ij}-\max\{r_i,r_j\})}}   &, d_{ij}\le \max\{r_i,r_j\}\\
      \beta_{ij}  &, d_{ij}> \max\{r_i,r_j\}
    \end{aligned}
    \right.
    \label{eq:model}
\end{equation}
where $d_{ij}$ is the structural proximity between node $i,j$, and $\beta_{i,j}$ is the feature proximity. $r_i$ and $r_j$ are connecting threshold parameters for node $i$ and $j$. The model illustrates that when two nodes have close structure proximity, the likelihood of connection depends only on the structure proximity. Otherwise, the likelihood of connection only depends on the feature proximity.

Previous studies~\citep{linkmoe,mao2023revisiting} show that MLPs primarily capture feature proximity,  but exhibit low correlation with structure proximity, an intuitive result since MLPs rely solely on node features. 
In contrast, heuristic methods primarily capture structure proximity~\citep{mao2023revisiting}. As a result, heuristic methods identify positive node pairs that MLPs would typically miss. Moreover, because each heuristic encodes a specific structural pattern, it is easier for an MLP to learn how these patterns relate to node features, thus improving performance beyond what feature proximity alone allows.
\begin{wrapfigure}{r}{0.48\textwidth}
  \centering
  \includegraphics[width=0.48\textwidth]{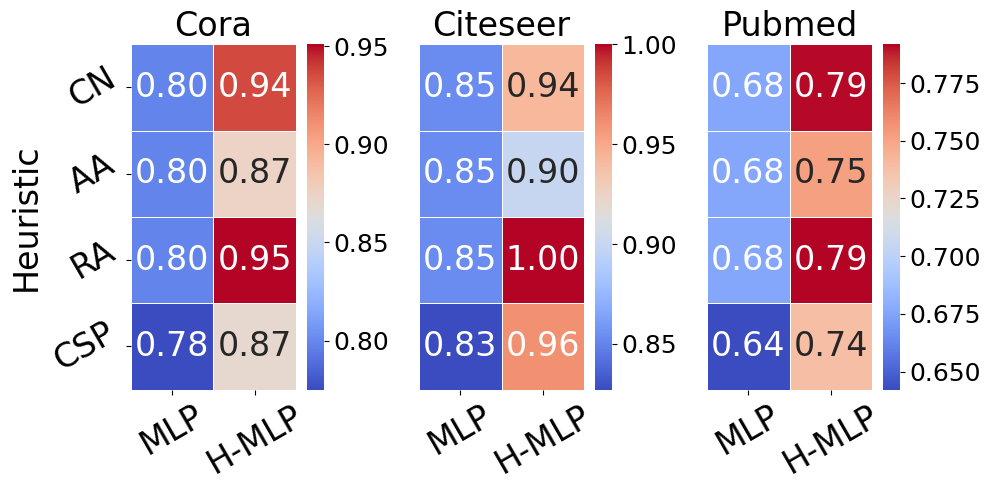}
  \caption{Subset ratio of positive edges identified by heuristic methods that are also recognized by MLPs (without distillation) and heuristic-distilled MLPs (H-MLP) across three datasets. The ratio indicates the proportion of positive edges identified by heuristic methods that are also recognized by MLPs.} 
    \label{fig:heatmap_complementary}
\end{wrapfigure}
To validate our claim, we compare the positive edges identified by heuristic methods, MLPs, and heuristic-distilled MLPs using the Hits@K protocol from~\citep{linkmoe} (K=5 for Citeseer, K=10 for Cora and Pubmed). A positive edge is selected if it ranks among the top-K predictions when compared to sampled negatives. We then compute the subset ratio, the proportion of heuristic-identified positives also captured by the MLP. As shown in Figure \ref{fig:heatmap_complementary}, this ratio increases significantly after distillation, indicating that heuristic-distilled MLPs effectively capture extra positive edges and learn complementary information.

Another important question is: \emph{why are heuristic methods better teachers than GNNs?} Our key insight is that although GNNs might have better accuracy, their predictions often rely heavily on structural information that is inaccessible to the student. As a result, the teachable knowledge they offer may be worse than that of heuristic methods. 

Formally, let $F(\bm{x}_i, \bm{x}_j, s_i, s_j)$ be a teacher model that approximates the ground-truth distribution $P(Y_{ij}=1)=p(\bm{x}_i, \bm{x}_j,s_i,s_j)$. 
The teachable knowledge of $F$ is defined as $\mathbb{E}F=\mathbb{E}_{s_i,s_j}\left(F(\bm{x}_i, \bm{x}_j,s_i,s_j)|\bm{x}_i, \bm{x}_j\right)$. 
Consider the KL Divergence
\begin{equation}
\begin{aligned}
    KL(p\|\mathbb{E}F)
    =
\underbrace{KL(p\|F)}_{\text{teacher error}}
    +\underbrace{\mathbb{E}_{\bm{x}_i, \bm{x}_j,s_i,s_j}\mathbb{E}_p\left(\log\mathbb{E}F-\log F\right)}_{\text{information lost}}
\label{eq:kl_decomposition}
\end{aligned}    
\end{equation}


Eq. \ref{eq:kl_decomposition} shows that the error of teachable knowledge consists of (1)
The teacher's error;
(2)  the information loss incurred when compressing the teacher’s predictions into a form that depends only on node features.
Even if a GNN has lower teacher error, the second term may be large if its predictions depend heavily on structural variables. In contrast, a heuristic that produces simpler, more feature-aligned predictions can yield lower total KL divergence after distillation.

We now present a toy example to concretely illustrate this phenomenon. Let all random variables $x_i,x_j,s_i,s_j$ all be binary. And $Y_{ij}=1$ if and only if $s_i=s_j=1$. Let $x_i,x_j{\sim}\text{Bern}(0.8)$, and let  the structural variables follow a Bernoulli distribution conditioned on features: 
\begin{equation}
    p_x:=P(s=1|x)=\left\{
    \begin{aligned}
        &0.5,\quad x_i=x_j=1,\\
        &0.6.\quad \text{otherwise},
    \end{aligned}
    \right.
\end{equation}
We compare two teacher models: 
\begin{compactenum}[\textbullet]
    \item  $F_1(x_i,x_j,s_i,s_j)=1$ if and only if $s_i=s_j=1$, and $0.1$ otherwise. The teachable knowledge is  $\mathbb{E}[F_1|x_i,x_j]=0.1+0.9p_x^2$.
    \item $F_2(x_i,x_j,s_i,s_j)=0.3$ (a constant predictor). The teachable knowledge is also a constant $0.3$.
\end{compactenum}
By computing the expected negative log-likelihood (detailed in Appendix \ref{app:example}), we find
\begin{equation}
\begin{aligned}
\mathbb{E}_{x, s}\left[\mathrm{NLL}(F_1)\right] &\approx 0.07 \quad &&\text{($F_1$ effectiveness)} \\
\mathbb{E}_{x, s}\left[\mathrm{NLL}(F_2)\right] &\approx 0.60 \quad &&\text{($F_1$ effectiveness)} \\
\mathbb{E}_{x}\left[\mathrm{NLL}(\mathbb{E}[F_1 \mid x])\right] &\approx 0.61 \quad &&\text{(Teachable knowledge of $F_1$ effectiveness)} \\
\mathbb{E}_{x}\left[\mathrm{NLL}(\mathbb{E}[F_2 \mid x])\right] &\approx 0.60 \quad &&\text{(Teachable knowledge of $F_2$ effectiveness)}
\end{aligned}
\end{equation} 

Although $F_1$ achieves high accuracy, its teachable knowledge is no better than that of a constant teacher, due to its strong dependence on structural variables that are inaccessible to the student. This highlights the central point: \emph{a model with better overall accuracy can still be a worse teacher if its predictions are not well aligned with the student’s input space.} Simpler models like heuristics may offer better guidance for feature-only students like MLPs.

\section{Ensemble Heuristic-Distilled MLPs}


Section \ref{sec:heuristic_good} highlights the advantages of using heuristics as teachers. However, since each heuristic captures only a specific type of structural proximity~\citep{mao2023revisiting}, an MLP distilled from a single heuristic is inherently limited. To improve effectiveness, we aim to enable MLPs to leverage multiple heuristics.

\begin{wrapfigure}{r}{0.4\textwidth}
    \centering
    \includegraphics[width=\linewidth,height=0.1\textheight]{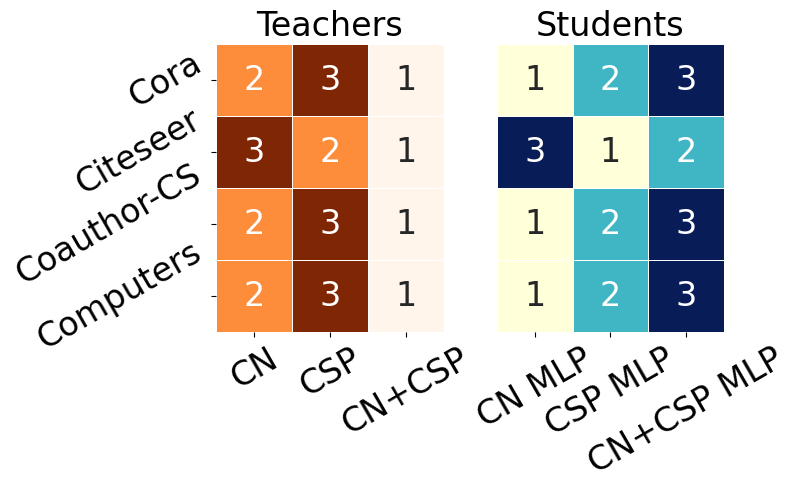}
    \caption{Accuracy rankings (Hits@20) of three teachers (CN, CSP, CN+CSP) and corresponding student MLPs across four datasets.}
    \label{fig:mix_heuristic_ranking}
    \vspace{-3mm}
\end{wrapfigure}

Ma et al.~\citep{linkmoe} demonstrated that combining multiple heuristics generally improves performance. So a straightforward approach is to ensemble heuristics and distill a single MLP from the combined signals. However, as discussed in Section \ref{sec:better_model}, stronger models do not always yield better students. Unfortunately, this holds true for mixing heuristics.
Figure~\ref{fig:mix_heuristic_ranking} shows that while CN+CSP outperforms CN and CSP individually, its distilled MLP performs worse. We hypothesize that mixing heuristics increases structural complexity, making it harder for the MLP to learn. These results suggest that direct distillation from combined heuristics is not effective.


\begin{wrapfigure}{r}{0.45\textwidth}
    \centering
    \begin{subfigure}{0.14\textwidth}
        \centering
        \includegraphics[width=\textwidth]{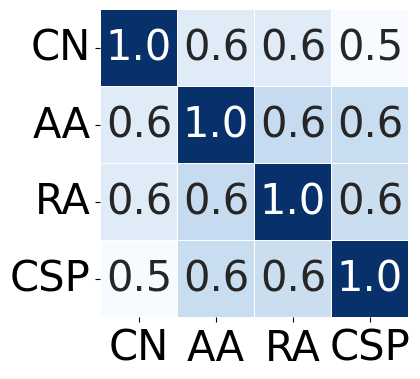}
        \caption{Cora}
        \label{fig:overlap_cora}
    \end{subfigure}
    \hfill
    \begin{subfigure}{0.14\textwidth}
        \centering
        \includegraphics[width=\textwidth]{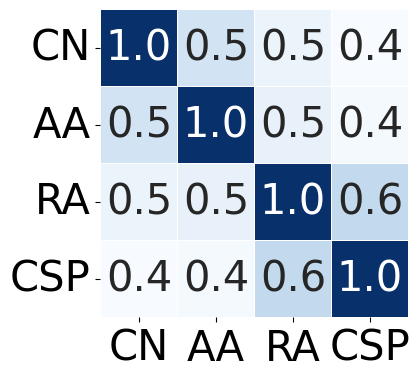}
        \caption{Pubmed}
        \label{fig:overlap_pubmed}
    \end{subfigure}
    \hfill
    \begin{subfigure}{0.14\textwidth}
        \centering
        \includegraphics[width=\textwidth]{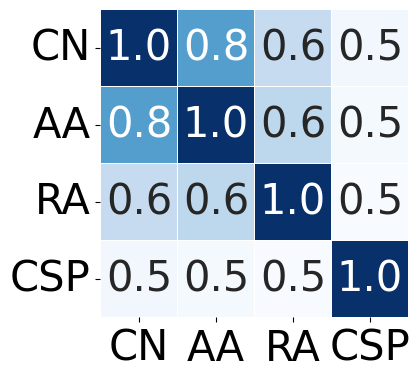}
        \caption{CS}
        \label{fig:overlap_cs}
    \end{subfigure}
    \caption{Overlap ratio of student MLPs distilled from different heuristic methods.}
    \label{fig:overlap}
\end{wrapfigure}

An alternative approach is to ensemble multiple MLPs. 
To effectively and efficiently ensemble multiple MLPs, our key insight is that: \emph{Since different heuristics capture different sets of positive links, their distilled MLPs should also capture different sets of positive links.} 
To validate this, we compute the overlapping ratio of positive edges identified by different distilled MLPs. We use Hits@10 to define the positive edge set for each distilled MLP. More details about positive edge set can be found in Section \ref{sec:heuristic_good}. The overlapping ratio is then computed as the Jaccard index between two positive edge sets. As shown in Figure \ref{fig:overlap}, the overlap among different heuristic-distilled MLPs is relatively low, confirming our hypothesis that each MLP captures complementary link information.

So a simple yet effective solution is to train a gating function that dynamically selects which heuristic-distilled MLP to use for each input node pair. Ma et al.~\citep{linkmoe} introduced a similar idea by training a gating network to ensemble different GNN4LP models. However, their gating network is not suitable for our setting, as it takes heuristic values as inputs, requiring neighbor-fetching operations, which would compromise the inference efficiency of MLPs.

To address this, we propose a gating MLP network that takes only the feature vectors of the target node pairs as input and outputs a weight for each distilled MLP. Our rationale is that node features alone should be sufficient for the gating function, since GNN-to-MLP distillation only works when node features correlate with structural information. The gating function is expected to learn which heuristic-derived MLP should be used based on the input node features.

\begin{wrapfigure}{r}{0.4\textwidth}
    \centering
    \includegraphics[width=0.9\linewidth]{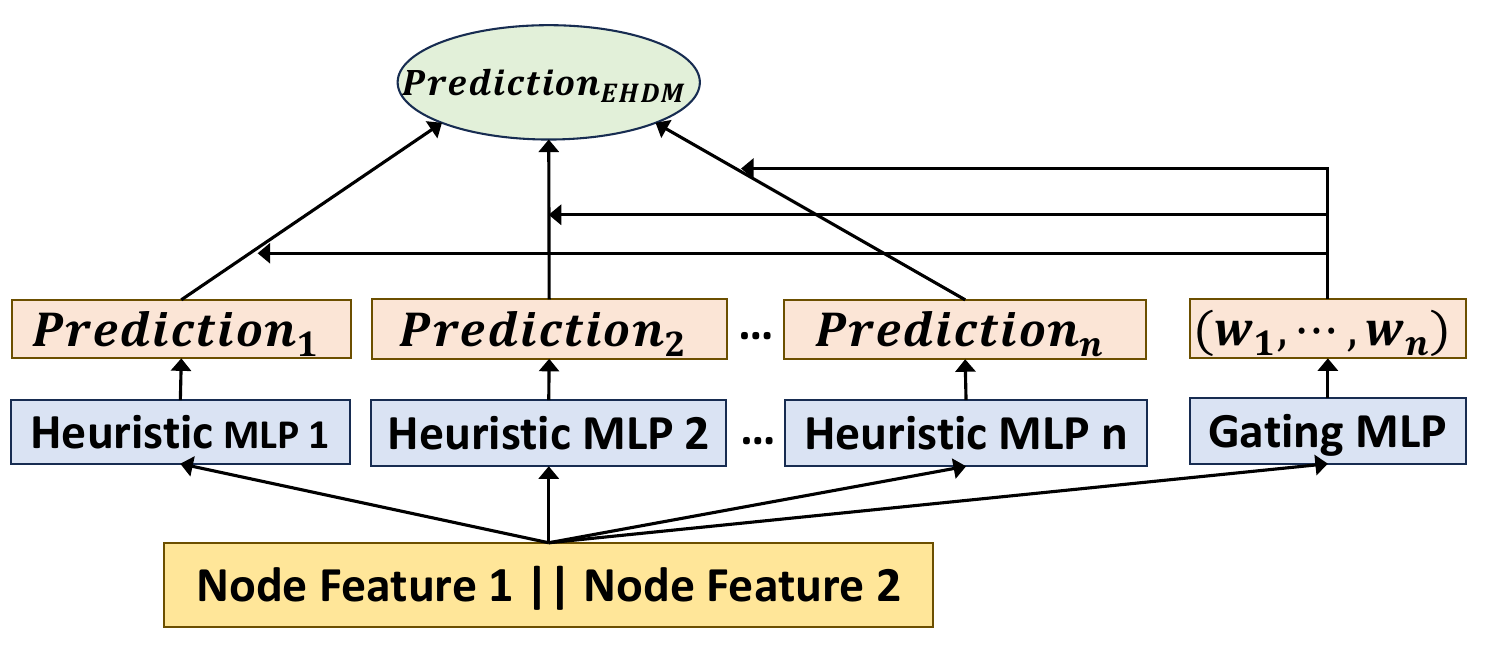}
    \caption{Illustration of Ensemble Heuristic-Distilled MLPs.}
    \label{fig:ensemble}
\end{wrapfigure}

The overall architecture is illustrated in Figure \ref{fig:ensemble}. Given an input node pair, the distilled MLPs and the gating MLP operate in parallel. The final prediction is then computed as a weighted sum of the predictions from the different MLPs. This approach ensures that: (1) No additional neighborhood fetching operations are required; (2) All MLPs can run simultaneously, minimizing inference overhead.


To train the gating function, we fix the parameters of the distilled MLPs and optimize the parameters of the gating function using the BCE loss. Additionally, we incorporate L1-regularization on the gate weights to encourage sparsity and improve generalization. The final loss function for training the gating MLP is:

\begin{equation}
    -y_{ij}\log(\sum_{h\in\mathcal{H}}w_hq^{(h)}_{i,j})-(1-y_{ij})\log(1-\sum_{h\in\mathcal{H}}w_hq^{(h)}_{i,j}) + \lambda \sum_{h\in\mathcal{H}}|w_h|
    \label{eq:loss_ensemble}
\end{equation}
where $y_{ij}$ is the ground-truth label, $\mathcal{H}$ is the heuristic set, and $q^{(h)}$ is the prediction of the MLP distilled from heuristic $h$, $w_h$ is the output weight of the gating function, and $\lambda$ is a hyper-parameter controlling the weight of L1-regularization.

Compared to other ensemble methods for link prediction~\citep{linkmoe}, the core novelty of EHDM is an efficient strategy to obtain a set of diverse and effective MLPs that complement each other by capturing different types of positive links. Simply training multiple MLPs via bootstrapping or random initialization is insufficient, as individual MLPs trained directly tend to perform poorly and focus mainly on feature proximity, capturing similar sets of positive links. And training multiple MLPs via boosting algorithms will increase the total training time, as they cannot be trained in parallel.

\section{Experiments}\label{sec:experiment}

Following previous works~\citep{guo2023linkless}, we include ten datasets with rich node features for evaluation. Dataset statistics are summarized in Table \ref{tab:dataset_stat} in the Appendix. 
For each dataset, we train three teacher GNNs: GCN~\citep{kipf2016semi}, SAGE~\citep{hamilton2017inductive}, and GAT~\citep{velivckovic2017graph}, and report the highest-performing one, denoted as ``\emph{Best GNN}''. Similarly, for the LLP baseline~\citep{guo2023linkless}, we distill three MLPs from GCN, SAGE, and GAT, respectively, and report the best-performing one, denoted as ``\emph{Best LLP}''. 
We include four heuristic methods in the experiments: CN~\citep{newman2001clustering}, AA~\citep{adamic2003friends}, RA~\citep{zhou2009predicting}, and CSP~\citep{liben2003link}.
\begin{table*}[ht]
    \centering
    \scriptsize
    \caption{Performance of GNN teacher and heuristic-based teachers across different datasets. 
    For Collab we report Hits@50, for IGB we report Hits@100, for Citation2 we report MRR, 
    for other datasets we report Hits@20. 
    \textbf{Bold} numbers mark the best performance in each row. 
    \underline{Underlined} numbers represent the 2nd best performance in each row.}
    \label{tab:teacher_performance}
    \begin{tabular}{@{}lll>{\columncolor[gray]{0.9}}c>{\columncolor[gray]{0.9}}c>{\columncolor[gray]{0.9}}c>{\columncolor[gray]{0.9}}c}
        \toprule
        \textbf{Dataset} & \textbf{MLP} & \textbf{Best GNN} & \textbf{CN} & \textbf{AA} & \textbf{RA} & \textbf{CSP} \\
        \midrule
        \textbf{Cora} & 
        \underline{73.21$\pm$3.28} & 
        \textbf{73.96$\pm$1.23} & 
       42.69$\pm$0.00 & 
       42.69$\pm$0.00 & 
       42.69$\pm$0.00 & 
       42.69$\pm$0.00 \\
        
        \textbf{Citeseer} & 
        \underline{68.26$\pm$1.92} & 
        \textbf{85.10$\pm$2.25} & 
       35.16$\pm$0.00 & 
       35.16$\pm$0.00 & 
       35.16$\pm$0.00 & 
       58.02$\pm$0.00 \\
        
        \textbf{Pubmed} & 
        \underline{49.40$\pm$3.53} & 
        \textbf{68.83$\pm$2.84} & 
       27.93$\pm$0.00 & 
       27.93$\pm$0.00 & 
       27.93$\pm$0.00 & 
       27.93$\pm$0.00 \\
        
        \textbf{CS} & 
       39.36$\pm$0.99 & 
       64.24$\pm$3.52 & 
       53.65$\pm$0.00 & 
        \textbf{74.32$\pm$0.00} & 
        \underline{74.01$\pm$0.00} & 
       0.00$\pm$0.00 \\
        
        \textbf{Physics} & 
       21.60$\pm$3.09 & 
       65.85$\pm$2.82 & 
       61.40$\pm$0.00 & 
        \underline{78.10$\pm$0.00} & 
        \textbf{79.80$\pm$0.00} & 
       0.00$\pm$0.00 \\
        
        \textbf{Computers} & 
       17.53$\pm$1.25 & 
       27.07$\pm$4.45 & 
       20.38$\pm$0.00 & 
        \underline{27.54$\pm$0.00} & 
        \textbf{32.67$\pm$0.00} & 
       0.00$\pm$0.00 \\
        
        \textbf{Photos} & 
       31.81$\pm$2.82 & 
        \textbf{49.79$\pm$8.87} & 
       34.37$\pm$0.00 & 
       42.63$\pm$0.00 & 
        \underline{45.39$\pm$0.00} & 
       0.00$\pm$0.00 \\
        
        \textbf{Collab} & 
       44.38$\pm$3.47 & 
       59.14$\pm$1.64 & 
       61.37$\pm$0.00 & 
        \textbf{64.17$\pm$0.00} & 
        \underline{63.81$\pm$0.00} & 
       46.49$\pm$0.00 \\
        
        \textbf{IGB} & 
        \underline{19.13$\pm$1.34} & 
        \textbf{20.47$\pm$1.39} & 
       7.78$\pm$0.00 & 
       7.78$\pm$0.00 & 
       7.78$\pm$0.00 & 
       1.00$\pm$0.00 \\
        
        \textbf{Citation2} & 
       39.17$\pm$0.44 & 
        \textbf{84.90$\pm$0.06} & 
       74.30$\pm$0.00 & 
       75.96$\pm$0.00 & 
        \underline{76.04$\pm$0.00} & 
       0.28$\pm$0.00 \\
        \bottomrule
    \end{tabular}
\end{table*}

\begin{table*}[ht]
    \centering
    \scriptsize
    \caption{Performance of MLP students distilled from GNN and heuristic-based methods across different datasets. For Collab we report Hits@50, for IGB we report Hits@100, for Citation2 we report MRR, for other datasets we report Hits@20. 
    \textbf{Bold} numbers mark the best performance, 
    \underline{underlined} numbers mark the 2nd best performance.}
    \label{tab:student_performance}
    \begin{tabular}{@{}l l l >{\columncolor[gray]{0.9}}l >{\columncolor[gray]{0.9}}l >{\columncolor[gray]{0.9}}l >{\columncolor[gray]{0.9}}l}
    \toprule
    \textbf{Dataset} & \textbf{MLP} & \textbf{Best LLP} & \textbf{CN MLP} & \textbf{AA MLP} & \textbf{RA MLP} & \textbf{CSP MLP} \\
    \midrule

    \textbf{Cora} & 
   73.21$\pm$3.28 & 
    \textbf{76.62$\pm$2.00} & 
   75.75$\pm$2.45 & 
   74.84$\pm$3.48 & 
    \underline{75.94$\pm$4.35} & 
   74.16$\pm$4.52 \\

    \textbf{Citeseer} & 
   68.26$\pm$1.92 & 
    \underline{76.53$\pm$4.88} & 
   75.69$\pm$1.14 & 
   70.90$\pm$2.24 & 
   75.38$\pm$1.36 & 
    \textbf{77.10$\pm$2.10} \\

    \textbf{Pubmed} & 
   49.40$\pm$3.53 & 
    \underline{59.95$\pm$1.58} & 
   56.98$\pm$2.47 & 
   54.95$\pm$2.91 & 
    \textbf{60.21$\pm$3.71} & 
   57.82$\pm$3.36 \\

    \textbf{CS} & 
   39.36$\pm$0.99 & 
   66.64$\pm$1.80 & 
   67.53$\pm$5.26 & 
    \textbf{69.56$\pm$2.97} & 
    \underline{69.39$\pm$2.97} & 
   67.41$\pm$1.34 \\

    \textbf{Physics} & 
   21.60$\pm$3.09 & 
    \underline{60.19$\pm$2.93} & 
   57.09$\pm$4.13 & 
    \textbf{61.10$\pm$3.15} & 
   52.87$\pm$4.12 & 
   50.80$\pm$2.09 \\

    \textbf{Computers} & 
   17.53$\pm$1.25 & 
   25.80$\pm$4.80 & 
    \underline{27.55$\pm$2.88} & 
    \textbf{30.45$\pm$4.27} & 
   25.50$\pm$1.97 & 
   21.97$\pm$1.26 \\

    \textbf{Photos} & 
   31.81$\pm$2.82 & 
    \underline{39.66$\pm$2.94} & 
   39.17$\pm$4.18 & 
    \textbf{40.23$\pm$4.07} & 
   30.84$\pm$3.25 & 
   38.49$\pm$5.06 \\

    \textbf{Collab} & 
   44.38$\pm$3.47 & 
    \textbf{49.30$\pm$0.79} & 
   48.23$\pm$0.89 & 
   47.33$\pm$1.02 & 
   48.93$\pm$0.66 & 
    \underline{48.99$\pm$0.69} \\

    \textbf{IGB} & 
   19.13$\pm$1.34 & 
    \textbf{25.12$\pm$1.14} & 
   24.38$\pm$0.11 & 
   24.20$\pm$0.90 & 
   24.11$\pm$1.25 & 
    \underline{24.75$\pm$0.83} \\

    \textbf{Citation2} & 
   39.17$\pm$0.44 & 
   42.78$\pm$0.10 & 
   43.05$\pm$0.23 & 
    \textbf{43.30$\pm$0.08} & 
   42.90$\pm$0.12 & 
    \underline{43.17$\pm$0.11} \\

    \bottomrule
    \end{tabular}
\end{table*}

\begin{table*}[htbp]
    \centering
    \caption{Comparison between \method{} and baselines across datasets.
    $\Delta_{\text{LLP}}$ and $\Delta_{\text{GNN}}$ denote the relative improvement (in \%) of our ensemble method with respect to the best LLP and best GNN, respectively. 
    \textbf{Bold} numbers mark the best performance, and \underline{underline} numbers mark the 2nd best performance. For Collab we report Hits@50, for IGB we report Hits@100, for Citation2 we report MRR, for other datasets we report Hits@20. }
    \label{tab:ensemble}
    \scriptsize
\begin{tabular}{@{}lcccccrr@{}}
\toprule
\textbf{Dataset} & \textbf{MLP} & \textbf{Best GNN} & \textbf{Best LLP} && \textbf{\method{}} & $\Delta_{\text{GNN}}$ & $\Delta_{\text{LLP}}$ \\
\midrule
\textbf{Cora} & 
73.21$\pm$3.28 &
73.96$\pm$1.23 & 
\underline{76.62$\pm$2.00} && 
\textbf{80.49$\pm$1.51} & 
+8.83\% & 
+5.06\% \\

\textbf{Citeseer} & 
68.26$\pm$1.92 &
\textbf{85.10$\pm$2.25} & 
76.53$\pm$4.88 && 
\underline{79.08$\pm$1.90} & 
-7.08\% & 
+3.33\% \\

\textbf{Pubmed} & 
49.40$\pm$3.53 &
\textbf{68.83$\pm$2.84} & 
59.95$\pm$1.58 && 
\underline{60.75$\pm$3.42} & 
-11.74\% & 
+1.33\% \\

\textbf{CS} & 
39.36$\pm$0.99 &
64.24$\pm$3.52 & 
\underline{66.64$\pm$1.80} && 
\textbf{74.33$\pm$2.59} & 
+15.71\% & 
+11.53\% \\

\textbf{Physics} & 
21.60$\pm$3.09 &
\textbf{65.85$\pm$2.82} & 
60.19$\pm$2.93 && 
\underline{64.44$\pm$5.22} & 
-2.14\% & 
+7.06\% \\

\textbf{Computers} & 
17.53$\pm$1.25 &
\underline{27.07$\pm$4.45} & 
25.80$\pm$4.80 && 
\textbf{30.41$\pm$2.90} & 
+12.35\% & 
+17.89\% \\

\textbf{Photos} & 
31.81$\pm$2.82 &
\textbf{49.79$\pm$8.87} & 
39.66$\pm$2.94 && 
\underline{45.89$\pm$1.67} & 
-7.83\% & 
+15.71\% \\

\textbf{Collab} & 
44.38$\pm$3.47 &
\textbf{59.14$\pm$1.64} & 
\underline{49.30$\pm$0.79} && 
49.27$\pm$0.88 & 
-16.70\% & 
-0.07\% \\

\textbf{IGB} & 
17.74$\pm$1.20 &
20.47$\pm$0.82 & 
\underline{25.12$\pm$1.14} && 
\textbf{27.27$\pm$0.27} & 
+33.21\% & 
+8.54\% \\

\textbf{Citation2} & 
38.12$\pm$0.18 &
\textbf{84.90$\pm$0.06} & 
42.78$\pm$0.10 && 
\underline{46.58$\pm$0.10} & 
-45.13\% & 
+8.87\% \\
\bottomrule
\end{tabular}
\end{table*}

\subsection{How do Heuristic-Distillation Compare to GNN-Distillation}

First, we comprehensively evaluate the effectiveness of heuristic teachers for MLP distillation. Table \ref{tab:teacher_performance} compares the teacher performance, while Table \ref{tab:student_performance} reports the performance of their corresponding student models. We observe that despite heuristic methods having significantly lower performance on certain datasets, heuristic-distilled MLPs consistently match or surpass the performance of the best GNN-distilled MLPs. These results empirically demonstrate that heuristic methods can serve as effective teachers for training MLPs.

\subsection{Performance of Ensemble Approach}

\begin{wrapfigure}{r}{0.4\textwidth}
    \centering
    \includegraphics[width=\linewidth]{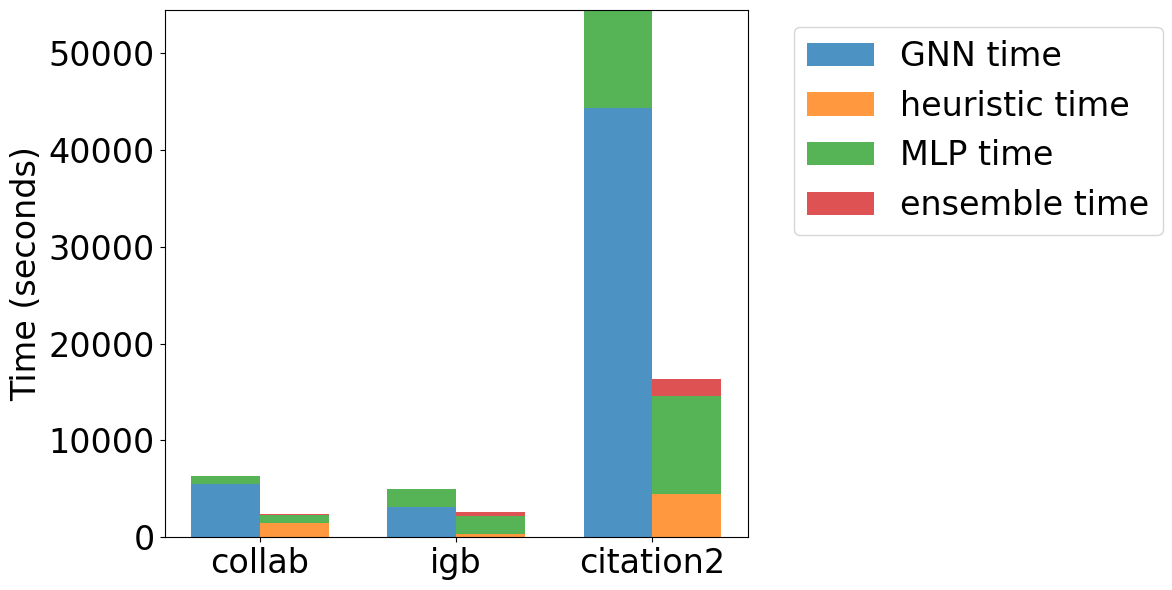}
    \caption{Training time decomposition of SAGE-LLP and \method{}.}
    \label{fig:ensemble_time}
    \vspace{-3mm}
\end{wrapfigure}

Next, we evaluate the efficiency and effectiveness of our ensemble approach.
Figure \ref{fig:ensemble_time} illustrates the training time decomposition of our ensemble method compared to LLP. The \method{} pipeline consists of three stages: heuristic guidance computation, MLP distillation, and ensembling. Although heuristic guidance must be generated for multiple heuristics, and MLP distillation is performed separately for each heuristic, both of these stages run in parallel. Therefore, we take the maximum time among the four heuristics as the effective time cost for each of these stages. We observe that by eliminating GNN training, our ensemble approach accelerates the entire distillation process by 1.95–3.32$\times$ compared to LLP. We also show \method{} has similar inference time as LLP in Appendix \ref{app:extra_exp}.

Table \ref{tab:ensemble} compares the performance of MLP, GNN, LLP (GNN-distilled MLP), and \method{}. The results show that, despite less training time, \method{} outperforms both MLP and LLP by leveraging the strengths of different heuristic-distilled MLPs. It is on average $7.93\%$ better than the best LLP model. Notably, it achieves at least 90\% of the best GNN effectiveness on 7 out of 10 datasets and even surpasses the best GNN on 4 datasets. For the two datasets where \method{} and LLP perform suboptimally (i.e., \texttt{Collab} and \texttt{Citation2}), we attribute this to their simple node features, as both datasets use only 128-dimensional node embeddings, the smallest feature size among all datasets. 

\section{Conclusion}

In this paper, we demonstrate that simple heuristic methods, despite their lower accuracy, can be surprisingly effective teachers for MLPs, enabling competitive link prediction performance while drastically reducing training costs.
We provide both empirical and theoretical analysis to explain this observation.
Furthermore, we introduce an ensemble approach that aggregates multiple heuristic-distilled MLPs using a gating mechanism. Extensive experiments show that this approach substantially reduces training time while consistently improving prediction accuracy.
Our findings have important implications for real-world link prediction tasks, particularly in large-scale web applications.

\section{Acknowledgement}

This work was partially supported by NSF 2211557, NSF 2119643, NSF 2303037, NSF 2312501, NSF 2531008, SRC JUMP 2.0 Center, Amazon Research Awards, and Snapchat Gifts.

\bibliographystyle{unsrtnat}
\bibliography{uai2025-template}

\newpage



\appendix
\section{LLP framework\label{app:llp}}

LLP captures relational knowledge by defining an anchor node \( v \) and a context set \( C_v \), training the student MLP to learn the relative ranking of context nodes based on their connection probability with \( v \). Let \( p_{v,i} \) and \( q_{v,i} \) denote the teacher GNN’s and student MLP’s predicted probabilities of link \( (v,i) \), respectively. In addition to cross entropy loss with ground-truth edge labels, LLP employs two distillation losses:

\begin{compactitem}[\textbullet]
    \item \textbf{Ranking-Based Loss} (\(L_R\)), ensuring the student MLP preserves the teacher’s ranking order:
    \begin{equation}
    L_R = \sum_{v\in\mathcal{V}}\sum_{i,j\in C_v}\max(0, -r\cdot(q_{v,i}-q_{v,j})+\delta)
    \label{eq:LR}
    \end{equation}
    where \( r \) is defined as:
    \begin{equation}
    r=\left\{
    \begin{aligned}
        1&,\quad \text{if } p_{v,i}-p_{v,j} > \delta\\
        -1&,\quad \text{if } p_{v,i}-p_{v,j} < -\delta\\
        0&,\quad \text{otherwise}
    \end{aligned}
    \right.
    \end{equation}
    with hyperparameter \( \delta \) controlling ranking sensitivity.
    \item \textbf{Distribution-Based Loss} (\(L_D\)), aligning student MLP predictions with the teacher GNN with temperature hyperparameter \( t \): 
\end{compactitem}
\begin{equation}
    L_D = \sum_{v\in \mathcal{V}}\sum_{i\in C_v}\frac{\exp(\frac{p_{v,i}}{t})}{\sum_{j\in C_v}\exp(\frac{p_{v,j}}{t})}
    \log\frac{exp(\frac{q_{v,i}}{t})}{\sum_{j\in C_v}\exp(\frac{q_{v,j}}{t})}
\label{eq:LD}
\end{equation}

\section{Proof of Lemma \ref{lemma}} \label{app:lemma}

\begin{proof}
Expanding the first objective using the definition of KL divergence:
\[
\begin{aligned}
\mathbb{E}_{x_i, x_j, s_i, s_j} \left[ \mathrm{KL}(F \,\|\, g) \right]\\
= &\mathbb{E}_{x_i, x_j, s_i, s_j} \left[ \sum_y F(y \mid x_i, x_j, s_i, s_j) \log \frac{F(y \mid x_i, x_j, s_i, s_j)}{g(y \mid x_i, x_j)} \right] \\
=& \mathbb{E}_{x_i, x_j} \left[ \sum_y \left( \mathbb{E}_{s_i, s_j \mid x_i, x_j} [F(y \mid x_i, x_j, s_i, s_j)] \cdot \log \frac{ \mathbb{E}_{s_i, s_j \mid x_i, x_j}[F(y \mid \cdot)] }{g(y \mid x_i, x_j)} \right) \right] + C \\
=& \mathbb{E}_{x_i, x_j} \left[ \mathrm{KL}\left( \mathbb{E}[{F}|x_i,x_j] \,\|\, g(y \mid x_i, x_j) \right) \right] + C,
\end{aligned}
\]
where $C$ is a constant independent of $g$, arising from the entropy terms of $F$. 

Meanwhile, since $\mathbb{E}[F|x_i,x_j]$ are $g$ are both independent to $s_i,s_j$, so
$$
\mathbb{E}_{x_i,x_j,s_i,s_j}KL(\mathbb{E}[F|x_i,x_j],g)=\mathbb{E}_{x_i,x_j}KL(\mathbb{E}[F|x_i,x_j],g)
$$
Therefore, the two objectives are equivalent.
\end{proof}

\section{Proof of Theorem \ref{thm:distillation-parity}}\label{app:proof}

\begin{proof}
    The teachable knowledge of $F_\text{GNN4LP-MLP}$ is 
    \begin{equation}
        \mathbb{E}_{s^\text{GNN}, s^\text{extra}} \left[ F_\text{GNN4LP} \mid \bm{x}_i, \bm{x}_j \right]
    \end{equation}
    Let a GNN function be 
    \[
    F_\text{GNN}(\bm{x}_i, \bm{x}_j, s_i^\text{GNN}, s_j^\text{GNN}) = \mathbb{E}[F_{GNN4LP}|\bm{x}_i, \bm{x}_j,s_i,s_j]
    \]
    Then its teachable knowledge 
    \[
     \mathbb{E}[F_{GNN}|\bm{x}_i, \bm{x}_j]= \mathbb{E}\{\mathbb{E}[F_{GNN4LP}|\bm{x}_i, \bm{x}_j,s_i,s_j] \mid \bm{x}_i, \bm{x}_j \}
    \]
    With the tower property of the expectation, i.e. $\mathbb{E}(f|X)=\mathbb{E}(\mathbb{E}(f|X,Y)|X)$, we know that 
    \begin{equation}
          \mathbb{E}[F_{GNN}|\bm{x}_i, \bm{x}_j]=\mathbb{E}[F_{GNN4LP}|\bm{x}_i, \bm{x}_j]
    \end{equation}
    Therefore we prove that for any GNN4LP we can find a GNN teacher to have the same teachable knowledge.
\end{proof}

\nop{
\begin{proof}
By assumption, $s_i^\text{extra} \perp\!\!\!\perp \bm{x}_i$. Thus,
    \[
    P(s_i^\text{extra} \mid \bm{x}_i) = P(s_i^\text{extra}).
    \]
    So 
    \[
    F_\text{GNN4LP-MLP}^*(\bm{x}_i, \bm{x}_j) = \mathbb{E}_{s^\text{GNN}, s^\text{extra}} \left[ F_\text{GNN4LP} \mid \bm{x}_i, \bm{x}_j \right]
    =\int F_\text{GNN4LP} \, dP(s^\text{GNN} \mid \bm{x}_i, \bm{x}_j) dP(s^\text{extra})
    \]
    Define a GNN teacher:
    \[
    F_\text{GNN}(\bm{x}_i, \bm{x}_j, s_i^\text{GNN}, s_j^\text{GNN}) = \mathbb{E}[F_{GNN4LP}|x_i,x_j,s_i,s_j] = \int F_\text{GNN4LP} \, dP(s^\text{extra})
    \]
    This GNN marginalizes over $s^\text{extra}$, which is independent of $\bm{x}_i, \bm{x}_j$.
    
    The optimal MLP distilled from $F_\text{GNN}$ is:
    \[
    \mathbb{E}_{s^\text{GNN}} \left[ F_\text{GNN} \mid \bm{x}_i, \bm{x}_j \right] = \mathbb{E}_{s^\text{GNN}, s^\text{extra}} \left[ F_\text{GNN4LP} \mid \bm{x}_i, \bm{x}_j \right] = F_\text{GNN4LP-MLP}^*.
    \]
    Thus, the teachable knowledge from $F_\text{GNN4LP}$ and $F_\text{GNN}$ are identical.
\end{proof}
}

\section{Computation Steps of Toy Example}\label{app:example}

Let $x_i,x_j,s_i,s_j$ all be binary random variables. And $Y_{ij}=1$ if and only if $s_i=s_j=1$. Let $x_i,x_j\overset{iid}{\sim}\text{Bern}(0.8)$. Let $s|x\sim\text{Bern}(p_x)$, with
\begin{equation}
    p_x=\left\{
    \begin{aligned}
        &0.5,\quad x_i=x_j=1,\\
        &0.6.\quad \text{otherwise},
    \end{aligned}
    \right.
\end{equation}
Let us consider two teacher functions. The first teacher $F_1(x_i,x_j,s_i,s_j)=1$ if and only if $s_i=s_j=1$, and $F_1=0.1$ otherwise. The second teacher $F_2$ is a constant predictor and $F_2=0.3$. So the teachable knowledge of $F_1$ is $\mathbb{E}[F_1|x_i,x_j]=0.1+0.9p_x^2$. And the teachable knowledge of $F_2$ is $\mathbb{E}[F_2|x_,x_j]=0.3$.

For a Bernoulli$(p_x^2)$ target and a predictor $q$, the conditional expected NLL is
\[
\mathbb{E}\bigl[-\ln q(Y)\mid x\bigr]
= p_x^2\bigl(-\ln q\bigr)\;+\;(1-p_x^2)\bigl(-\ln(1-q)\bigr).
\]

The following computation steps are illustrated in Table \ref{tab:step4} and Table \ref{tab:step5}.

\begin{table}[h]
\centering
\tiny
\caption{Conditional expected NLL given $x$. Here $p_x=0.5$ if $(x_i,x_j)=(1,1)$ (prob.~0.64), else $p_x=0.6$ (prob.~0.36).}\label{tab:step4}
\begin{tabular}{lllc}
\toprule
Region & Predictor & $q$ & $\mathbb{E}[-\ln q(Y)\mid x]$ \\
\midrule
\multirow{4}{*}{$(1,1)$, $p_x^2=0.25$}
  & $F_1$ & $q_1=1\,(Y=1),\,0.1\,(Y=0)$ & $(1-0.25)(-\ln0.9)\approx0.0790$ \\
  & $F_2$ & $q_2=0.3$                    & $0.25(-\ln0.3)+0.75(-\ln0.7)\approx0.5685$ \\
  & $\mathbb{E}F_1$ & $q_{g1}=0.1+0.9\cdot0.25=0.325$ & $\approx0.5758$ \\
  & $\mathbb{E}F_2$ & $q_{g2}=0.3$                    & $\approx0.5685$ \\
\midrule
\multirow{4}{*}{otherwise, $p_x^2=0.36$}
  & $F_1$ & $q_1=0.1$                     & $(1-0.36)(-\ln0.9)\approx0.0674$ \\
  & $F_2$ & $q_2=0.3$                     & $0.36(-\ln0.3)+0.64(-\ln0.7)\approx0.6617$ \\
  & $\mathbb{E}F_1$ & $q_{g1}=0.1+0.9\cdot0.36=0.424$ & $\approx0.6619$ \\
  & $\mathbb{E}F_2$ & $q_{g2}=0.3$                    & $\approx0.6617$ \\
\bottomrule
\end{tabular}
\end{table}

\begin{table}[h]
\centering
\caption{Unconditional expected NLL:
$\Pr((1,1))=0.64,\ \Pr(\text{otherwise})=0.36$.}\label{tab:step5}
\begin{tabular}{lc}
\toprule
Predictor   & $\displaystyle \mathbb{E}[-\ln q(Y)]$ \\
\midrule
$F_1$  & $0.64\cdot0.0790 + 0.36\cdot0.0674 \approx 0.0748$ \\
$F_2$   & $0.64\cdot0.5685 + 0.36\cdot0.6617 \approx 0.6021$ \\
$\mathbb{E}F_1$   & $0.64\cdot0.5758 + 0.36\cdot0.6619 \approx 0.6068$ \\
$\mathbb{E}F_2$   & same as Teacher 2: $\approx0.6021$ \\
\bottomrule
\end{tabular}
\end{table}

\section{Experiment Setting}
\textbf{Datasets}. Following previous works~\citep{guo2023linkless}, we include ten datasets for a comprehensive evaluation, i.e., \texttt{Cora}, \texttt{Citeseer}, \texttt{Pubmed}~\citep{yang2016revisiting}, \texttt{Coauthor-CS}, \texttt{Coauthor-Physics}~\citep{shchur2018pitfalls}, \texttt{Amazon-Computers}, \texttt{Amazon-photos}~\citep{mcauley2015image,shchur2018pitfalls}, \texttt{OGBL-Collab}, \texttt{OGBL-Citation2}~\citep{wang2020microsoft}, and \texttt{IGB}~\citep{khatua2023igb}.  Dataset statistics are summarized in Table \ref{tab:dataset_stat} in the appendix. Notably, \texttt{OGBL-Collab} contains over a million edges, while \texttt{IGB} and \texttt{OGBL-Citation2} have over a million nodes and tens of millions of edges. 


\textbf{Evaluation Protocol}. For \texttt{Collab} and \texttt{Citation2} datasets, we use the same train/test split as in the OGBL benchmark~\citep{ogb}, where the edges are split according to time to simulate real-world production settings. For all other datasets, we randomly sample 5\%/15\% of the edges along with an equal number of non-edge node pairs, as validation/test sets. Unlike LLP~\citep{guo2023linkless}, which trains the teacher GNN with ten different random seeds and selects the model with the highest validation accuracy, we train the teacher GNN only once for distillation. Our approach better reflects practical constraints, as training a GNN ten times is infeasible for real-world applications. 
For each dataset, we train three teacher GNNs: GCN~\citep{kipf2016semi}, SAGE~\citep{hamilton2017inductive}, and GAT~\citep{velivckovic2017graph}, and report the highest-performing one, denoted as ``\emph{Best GNN}''. Similarly, for the LLP baseline~\citep{guo2023linkless}, we distill three MLPs from GCN, SAGE, and GAT, respectively, and report the best-performing one, denoted as ``\emph{Best LLP}''. 
We include four heuristic methods: CN~\citep{newman2001clustering}, AA~\citep{adamic2003friends}, RA~\citep{zhou2009predicting}, and CSP~\citep{liben2003link}.
For each dataset, all MLPs have same architecture.
For Collab we report Hits@50, for IGB we report Hits@100, for Citation2 we report MRR, for other datasets we report Hits@20. 

\section{Hyper-parameters}

\begin{table}[ht]
    \centering
    \caption{Statistics of datasets.}
    \label{tab:dataset_stat}
    \begin{tabular}{@{}llll@{}}\toprule
      Dataset   & \# Nodes & \# Edges & \# Features \\\midrule
       Cora  & 2,708 & 5,278 & 1,433\\
       Citeseer & 3,327 & 4,552 & 3,703\\
       Pubmed & 19,717 & 44,324 & 500\\
       Coauthor-CS & 18,333 & 163,788 & 6,805\\
       Coauthor-Physics & 34,493 & 495,924 & 8,415\\
       Computers & 13,752 & 491,722 & 767\\
       Photos & 7,650 & 238,162 & 745\\
       OGBL-Collab & 235,868 & 1,285,465 & 128\\
       IGB & 1M & 12M & 1,024\\
       OGBL-Citation2 & 2.9M & 30.6M & 128\\
       \bottomrule
    \end{tabular}
\end{table}

For capped shortest path (CSP) heuristic, we set the upper bound $\tau=6$ for seven smaller datasets and $\tau=2$ for the three larger datasets.
When distilling MLPs from teacher models (GNNs or heuristic methods), we follow the same hyperparameter configurations as in~\citep{guo2023linkless} for sampling context nodes and adopt the configurations from~\citep{heart} for training teacher GNNs and non-distilled MLPs. 
We use 3-layer MLPs for \texttt{OGBL-Collab}, \texttt{IGB}, and \texttt{OGBL-Citation2}, and 2-layer MLPs for all other datasets. 
Let $\alpha$, $\beta$ denote the loss weights of $L_R$ and $L_D$, respectively. After distillation, we perform a grid search to determine the loss weights $\alpha$ and $\beta$, where $\alpha,\beta\in\{0,0.001,1,10\}$. Additionally, we use grid search to optimize the margin $\delta$ in $L_R$ (Eq. \ref{eq:LR}), where $\delta\in\{0.05,0.1,0.2\}$.
For training ensemble models, we conduct a grid search to find the optimal weight $\lambda$ for L1 regularization (Eq. \ref{eq:loss_ensemble}), where $\lambda \in \{0,0.1,1\}$.

\section{Additional Experiment Results}\label{app:extra_exp}

\subsection{Inference Speed}

One of the primary advantages of our approach is its fast inference training speed. Figure 7 in our paper has shown its advantage in training speed. And prior work, such as LLP, has demonstrated that GNN-to-MLP methods can achieve up to 70 times faster inference compared to GNNs. Our experimental results, as shown in Table \ref{tab:infer_speed}, confirm that EHDM maintains similar inference speed advantages.

\begin{table}[h]
\centering
\caption{Inference time (in ms) for different models on \texttt{Collab} dataset.}
\label{tab:infer_speed}
\begin{tabular}{@{}lccc@{}}
\toprule
Model & SAGE & SAGE-LLP & EHDM \\
\midrule
Inference Time (ms) & 134.3 & 2.1 & 3.6 \\
\bottomrule
\end{tabular}

\end{table}

\subsection{Breakdown of Best GNN and Best LLP}
Table \ref{tab:gnn_breakdown} and Table \ref{tab:llp_breakdown} shows the detailed breakdowns of ``best GNN'' and ``best LLP'', respectively. We observe that different datasets have different best GNNs and different best LLPs. So when compared with a fixed teacher GNN, our proposed method will have even better improvement than the number reported in the main paper.

\begin{table*}[ht]
\centering
\caption{Breakdown of teacher performance across datasets, showing GCN, SAGE, GAT, and the best-performing GNN. All values are given as mean $\pm$ standard deviation.}
\label{tab:gnn_breakdown}
\begin{tabular}{lcccc}
\toprule
\textbf{Dataset} & \textbf{GCN} & \textbf{SAGE} & \textbf{GAT} & \textbf{best GNN} \\
\midrule
\textbf{cora} & 
\(73.96 \pm 1.23\) & 
\(71.80 \pm 3.08\) & 
\(71.77 \pm 1.44\) & 
\(73.96 \pm 1.23\) \\

\textbf{citeseer} & 
\(79.91 \pm 2.52\) & 
\(85.10 \pm 2.25\) & 
\(80.48 \pm 2.21\) & 
\(85.10 \pm 2.25\) \\

\textbf{pubmed} & 
\(68.83 \pm 2.84\) & 
\(62.45 \pm 3.11\) & 
\(56.15 \pm 3.09\) & 
\(68.83 \pm 2.84\) \\

\textbf{cs} & 
\(64.24 \pm 3.52\) & 
\(51.20 \pm 2.61\) & 
\(59.17 \pm 4.15\) & 
\(64.24 \pm 3.52\) \\

\textbf{physics} & 
\(62.25 \pm 3.75\) & 
\(65.85 \pm 2.82\) & 
\(45.03 \pm 6.55\) & 
\(65.85 \pm 2.82\) \\

\textbf{computers} & 
\(27.07 \pm 4.45\) & 
\(26.00 \pm 3.09\) & 
\(9.82 \pm 1.82\) & 
\(27.07 \pm 4.45\) \\

\textbf{photos} & 
\(49.37 \pm 1.18\) & 
\(49.79 \pm 8.87\) & 
\(44.45 \pm 2.13\) & 
\(49.79 \pm 8.87\) \\

\textbf{collab} & 
\(56.75 \pm 1.39\) & 
\(59.14 \pm 1.64\) & 
\(55.90 \pm 1.22\) & 
\(59.14 \pm 1.64\) \\

\textbf{IGB} & 
\(20.47 \pm 0.82\) & 
\(20.38 \pm 1.39\) & 
\texttt{OOM} & 
\(20.47 \pm 0.82\) \\

\textbf{citation2} & 
\(84.90 \pm 0.06\) & 
\(82.92 \pm 0.22\) & 
\texttt{OOM} & 
\(84.90 \pm 0.06\) \\
\bottomrule
\end{tabular}
\end{table*}

\begin{table*}[ht]
\centering
\caption{Breakdown of student MLP performance distilled from GCN, SAGE, and GAT, along with the best-performing LLP (best LLP). All values are mean $\pm$ standard deviation.}
\label{tab:llp_breakdown}
\begin{tabular}{lcccc}
\toprule
\textbf{Dataset} & \textbf{GCN-LLP} & \textbf{SAGE-LLP} & \textbf{GAT-LLP} & \textbf{best LLP} \\
\midrule
\textbf{cora} & 
\(76.62 \pm 2.00\) & 
\(75.26 \pm 3.22\) & 
\(74.54 \pm 4.34\) & 
\(76.62 \pm 2.00\) \\

\textbf{citeseer} & 
\(76.53 \pm 4.88\) & 
\(73.27 \pm 4.11\) & 
\(74.95 \pm 3.14\) & 
\(76.53 \pm 4.88\) \\

\textbf{pubmed} & 
\(59.95 \pm 1.58\) & 
\(54.07 \pm 4.95\) & 
\(58.30 \pm 3.66\) & 
\(59.95 \pm 1.58\) \\

\textbf{cs} & 
\(66.64 \pm 1.80\) & 
\(65.50 \pm 3.91\) & 
\(63.65 \pm 1.97\) & 
\(66.64 \pm 1.80\) \\

\textbf{physics} & 
\(60.19 \pm 2.93\) & 
\(58.46 \pm 1.85\) & 
\(56.83 \pm 3.81\) & 
\(60.19 \pm 2.93\) \\

\textbf{computers} & 
\(23.66 \pm 1.61\) & 
\(25.80 \pm 4.80\) & 
\(23.51 \pm 2.14\) & 
\(25.80 \pm 4.80\) \\

\textbf{photos} & 
\(39.66 \pm 2.94\) & 
\(34.72 \pm 3.34\) & 
\(35.86 \pm 4.65\) & 
\(39.66 \pm 2.94\) \\

\textbf{collab} & 
\(49.30 \pm 0.79\) & 
\(47.99 \pm 0.62\) & 
\(48.45 \pm 0.82\) & 
\(49.30 \pm 0.79\) \\

\textbf{IGB} & 
\(25.12 \pm 1.14\) & 
\(24.33 \pm 0.48\) & 
\texttt{NA} & 
\(25.12 \pm 1.14\) \\

\textbf{citation2} & 
\(42.78 \pm 0.10\) & 
\(42.62 \pm 0.08\) & 
\texttt{NA} & 
\(42.78 \pm 0.10\) \\
\bottomrule
\end{tabular}
\end{table*}

\subsection{Comparision of GNN4LP and GNN methods}

Below we show a more comprehensive comparison between NCN and standard GNNs across seven datasets, including two million-scale graphs. As shown in the Table \ref{tab:app_teacher} and Table \ref{tab:app_student}, while NCN consistently achieves the highest accuracy as a teacher, the MLPs distilled from standard GNNs outperform those distilled from NCN in 6 out of 7 cases. Furthermore, our proposed EHDM model consistently achieves the best student performance, while also training 3 times faster.

\begin{table}[htbp]
\centering
\caption{Teacher Model Performance}
\label{tab:app_teacher}
\begin{tabular}{lcccc}
\toprule
Dataset & GCN & SAGE & GAT & NCN \\
\midrule
cora       & $73.96 \pm 1.23$ & $71.80 \pm 3.08$ & $71.77 \pm 1.44$ & $83.22 \pm 1.37$ \\
citeseer   & $79.91 \pm 2.52$ & $85.10 \pm 2.25$ & $80.48 \pm 2.21$ & $85.45 \pm 0.73$ \\
pubmed     & $68.83 \pm 2.84$ & $62.45 \pm 3.11$ & $56.15 \pm 3.09$ & $72.47 \pm 1.86$ \\
CS         & $64.24 \pm 3.52$ & $51.20 \pm 2.61$ & $59.17 \pm 4.15$ & $74.49 \pm 2.37$ \\
computers  & $27.07 \pm 4.45$ & $27.07 \pm 4.45$ & $26.00 \pm 3.09$ & $39.49 \pm 2.30$ \\
coauthor   & $56.75 \pm 1.39$ & $59.14 \pm 1.64$ & $55.90 \pm 1.22$ & $63.07$ \\
citation2  & $84.90 \pm 0.06$ & $82.92 \pm 0.22$ & OOM              & $89.11$ \\
\bottomrule
\end{tabular}
\end{table}

\begin{table}[htbp]
\centering
\caption{Distilled Model Performance}
\label{tab:app_student}
\begin{tabular}{lccccc}
\toprule
Dataset & GCN-MLP & SAGE-MLP & GAT-MLP & NCN-MLP & EHDM \\
\midrule
cora       & $76.62 \pm 2.00$ & $75.26 \pm 3.22$ & $74.54 \pm 4.34$ & $69.37 \pm 2.10$ & $80.49 \pm 1.51$ \\
citeseer   & $76.53 \pm 4.88$ & $73.27 \pm 4.11$ & $74.95 \pm 3.14$ & $79.29 \pm 2.79$ & $79.08 \pm 1.90$ \\
pubmed     & $59.95 \pm 1.58$ & $54.07 \pm 4.95$ & $58.30 \pm 3.66$ & $59.85 \pm 1.31$ & $60.75 \pm 3.42$ \\
CS         & $66.64 \pm 1.80$ & $65.50 \pm 3.91$ & $63.65 \pm 1.97$ & $58.82 \pm 4.01$ & $74.33 \pm 2.59$ \\
computers  & $23.66 \pm 1.61$ & $25.80 \pm 4.80$ & $23.51 \pm 2.14$ & $23.01 \pm 6.21$ & $30.41 \pm 2.90$ \\
collab     & $49.30 \pm 0.79$ & $48.00 \pm 0.62$ & $48.45 \pm 0.82$ & $48.00 \pm 1.79$ & $49.27 \pm 0.88$ \\
citation2  & $42.78 \pm 0.10$ & $42.62 \pm 0.08$ & NA               & $42.58 \pm 2.10$ & $46.58 \pm 0.10$ \\
\bottomrule
\end{tabular}
\end{table}

\subsection{Additional GNN4LP Teachers}

Table \ref{tab:more_gnn4lp} shows more performance of GNN4LP models and their distilled MLPs. Specifically, we evaluate NCN~\citep{ncnc}, NCNC~\citep{ncnc}, and BUDDY~\citep{buddy}. The results further validate our finding that stronger models might not be better teachers.

\begin{table*}[h]
    \centering
    \tiny
    \caption{Performance comparison (Hits@10) of different teacher models and their distilled MLP students across datasets.}
    \label{tab:more_gnn4lp}
    \begin{tabular}{l c c c c c c}
        \toprule
        & \multicolumn{3}{c}{Standard GNN} & \multicolumn{3}{c}{GNN4LP} \\
        \cmidrule(lr){2-4} \cmidrule(lr){5-7}
        Dataset & GCN & SAGE & GAT & NCN & NCNC & BUDDY \\
        \midrule
        \multicolumn{7}{l}{\textbf{Teacher Models}} \\
        \midrule
        Cora     & $66.11 \pm 4.93$ & $64.82 \pm 4.48$ & $61.97 \pm 6.47$ & $74.50 \pm 1.64$ & $76.59 \pm 3.58$ & $56.24 \pm 3.44$ \\
        Citeseer & $73.41 \pm 2.89$ & $77.45 \pm 3.60$ & $75.91 \pm 1.84$ & $80.18 \pm 1.76$ & $83.61 \pm 1.14$ & $79.34 \pm 2.03$ \\
        Pubmed   & $54.60 \pm 3.98$ & $44.70 \pm 6.03$ & $42.18 \pm 9.92$ & $61.02 \pm 3.59$ & $60.74 \pm 2.34$ & $48.25 \pm 4.30$ \\
        \midrule
        \multicolumn{7}{l}{\textbf{Student MLPs}} \\
        \midrule
        Cora     & $68.96 \pm 2.29$ & $64.29 \pm 8.34$ & $66.15 \pm 6.58$ & $57.53 \pm 5.95$ & $67.32 \pm 3.31$ & $65.09 \pm 3.36$ \\
        Citeseer & $68.22 \pm 3.18$ & $65.80 \pm 4.77$ & $70.68 \pm 2.92$ & $70.82 \pm 3.58$ & $72.91 \pm 1.72$ & $66.02 \pm 3.25$ \\
        Pubmed   & $47.62 \pm 3.22$ & $41.17 \pm 5.10$ & $43.89 \pm 5.16$ & $46.38 \pm 3.09$ & $46.73 \pm 4.52$ & $44.42 \pm 3.15$ \\
        \bottomrule
    \end{tabular}
\end{table*}

\subsection{Performance of Ensemble Heuristics}

Here, we provide more results showing distilling MLPs from ensemble heuristics is sub-optimal. We trained an MLP as a gating function to ensemble different heuristic methods and distilled MLPs from these ensemble heuristics. As shown in Table \ref{tab:heuristic_mlp_comparison}, although ensemble heuristics generally yield higher Hits@K scores, the student MLP distilled from the ensemble heuristics has lower Hits@K.
The results suggest that MLPs, limited by their capacity, struggle to effectively learn complex structural information. Consequently, GNNs and GNN4LP models, which integrate various proximities together \citep{mao2023revisiting,linkmoe}, may also present challenges for MLPs to fully capture.

\begin{table}[ht]
    \centering
    \caption{Performance comparison of different heuristic methods and their corresponding MLP-distilled versions across datasets. CC denotes ``CN+CSP'', CARC denotes ``CC+AA+RA+CSP''. \textbf{Bolded numbers} mark the highest performance. \underline{Underlined numbers} represent the second highest performance.}
    \label{tab:heuristic_mlp_comparison}
    \begin{tabular}{@{}lcccc@{}}
        \toprule
        Dataset & CN & CSP & CC & CARC \\
        \midrule
        Teacher & & & & \\
        Cora & 42.69 & 42.69 & \underline{60.34} & \textbf{60.76} \\
        Citeseer & 35.16 & 58.02 & \underline{59.21} & \textbf{69.10} \\
        Coauthor-CS & 53.65 & 0 & \underline{74.90} & \textbf{77.17} \\
        Computers & 20.38 & 0 & \underline{22.27} & \textbf{33.87} \\
        \midrule

        Student MLP & & & & \\
        Cora & \textbf{75.75} & \underline{74.16} & 69.60 & 69.49 \\
        Citeseer & 75.69 & \textbf{77.10} & \underline{75.74} & 72.26 \\
        Coauthor-CS & \textbf{67.53} & \underline{67.41} & 62.21 & 62.33 \\
        Computers & \textbf{27.55} & \underline{21.97} & 21.95 & 19.12 \\
        \bottomrule
    \end{tabular}
\end{table}

\subsection{Complicated Heuristic Teachers}

Notably, MLPs cannot learn well from Katz~\citep{katz1953new} heuristic either. Katz \citep{katz1953new} is a well-established heuristic for link prediction, widely effective across various datasets \citep{heart}. Katz is computed as $\sum_{l=1}^\infty \lambda^l |path^{(l)}(i,j)|$, where $\lambda<1$ is a damping factor and $|path^{(l)}(i,j)|$ is the number of length-$l$ paths between $i$ and $j$. As shown in Table \ref{tab:katz}, while Katz outperforms CN as a heuristic, the MLP distilled from Katz performs worse than MLPs trained without distillation. It suggests that MLPs cannot learn well from complicated heuristics.

\begin{table}[ht]
    \centering
    \footnotesize
    \caption{Hits@10 of CN, Katz, and MLP models across different datasets.}
    \label{tab:katz}
    \begin{tabular}{lcccccc}
        \toprule
        Dataset & CN & Katz & MLP & CN MLP & Katz MLP \\
        \midrule
        Cora & 42.69 & 51.61 & 58.48& 62.16  & 55.33  \\
        Citeseer & 35.16 & 57.36 & 61.21 & 70.51  & 66.64 \\
        Pubmed & 27.93 & 42.17 & 38.21 & 48.43 & 37.45 \\
        \bottomrule
    \end{tabular}
\end{table}

\subsection{Effects of the Number of Trainable Parameters}

To investigate whether the improvement of the ensemble model is solely due to an increase in trainable parameters, we conduct an ablation study. Specifically, we train an additional MLP, denoted as MLP*, with the same number of trainable parameters as the ensemble MLP. Table \ref{tab:ablation_num_param} compares the performance of MLP, MLP*, and our ensemble method. The results show that the ensemble method consistently outperforms MLP*, demonstrating that its effectiveness is not merely a result of increased parameter capacity.

\begin{table}[ht]
    \centering
    \caption{Performance comparison of MLP, MLP*, and Ensemble across different datasets.}
    \label{tab:ablation_num_param}
    \begin{tabular}{lccc}
        \toprule
        Dataset & MLP & MLP* & Ensemble \\
        \midrule
        Cora & 73.21 $\pm$ 3.28 & 71.05 $\pm$ 2.02 & 80.49 $\pm$ 1.51 \\
        Citeseer & 68.26 $\pm$ 1.92 & 68.57 $\pm$ 3.28 & 79.08 $\pm$ 1.90 \\
        Pubmed & 49.40 $\pm$ 3.53 & 57.10 $\pm$ 2.56 & 69.75 $\pm$ 3.42 \\
        IGB & 17.74 $\pm$ 1.20 & 20.92 $\pm$ 0.92 & 27.27 $\pm$ 0.27 \\
        \bottomrule
    \end{tabular}
\end{table}

\subsection{Problems with ``Production'' Setting}

Guo et al.~\citep{guo2023linkless} introduced a ``production'' setting for splitting train/test data in link prediction tasks, where a random subset of nodes is treated as newly added to the graph. These nodes and their associated edges are removed from the training graph. Their results showed a significant performance drop when transitioning from the ``transductive'' setting to the ``production'' setting. Similarly, we observe the same performance decline for heuristic-distilled MLPs, to the extent that heuristic distillation may fail to improve MLP performance.

Our analysis reveals that this performance drop is primarily due to the "production" split significantly altering the graph structure, leading to inaccurate heuristic values in the training graph. For instance, triangle counts play a crucial role in link prediction~\citep{adamic2003friends,zhang2021labeling,buddy}. As shown in Table \ref{tab:triangle}, the number of triangles in the training graphs is drastically reduced under the "production" setting compared to the ``transductive'' setting. This discrepancy indicates that the removing random nodes make the graphs more ``broken'' and that ``production'' setting introduces a substantial structural mismatch between training and test graphs, explaining the performance degradation observed in~\citep{guo2023linkless}.

In practice, such drastic structural changes are unlikely in real-world applications. Therefore, we argue that the ``production'' setting does not accurately reflect real-world scenarios. Instead, datasets like \texttt{Collab} and \texttt{Citation2}, which split edges based on time, offer a more realistic evaluation. Consequently, we do not adopt the ``production'' setting in our experiments.

\begin{table}[ht]
    \centering
    \footnotesize
    \caption{Triangle counts and the number of connected components (\#CC) in training graph under different split settings.}
    \label{tab:triangle}
    \begin{tabular}{lllll}
        \toprule
        & Cora & & Pubmed & \\
        & Triangle Counts & \#CC & Triangle Counts & \#CC\\
        \midrule
        Train graph (production) & 31 & 898 & 3,372 & 3,407 \\
        Train graph (transductive) & 991 & 170 & 7,633 & 1,512 \\
        Original graph & 1,630 & 78 & 12,520 & 1 \\
        \bottomrule
    \end{tabular}
    
\end{table}

\section{Limitations}

Since MLPs rely heavily on rich node features to perform effectively, our experiments focus primarily on datasets from social networks and recommendation systems, where such features are abundant. In contrast, biological graphs, such as protein-protein interaction networks, are not included in our evaluation, as they typically lack informative node features, making them less suitable for MLP-based approaches. This limitation is not specific to our method but is inherent to all MLP distillation methods on graphs.

\end{document}